\documentclass{article}
\usepackage{spconf,amsmath,graphicx,hyperref}
\usepackage[T1]{fontenc}
\usepackage{graphicx}
\usepackage{array}
\usepackage{siunitx} 
\usepackage{amssymb}  
\usepackage{setspace}
\usepackage{tabularx}
\usepackage{utfsym}
\usepackage{colortbl}
\usepackage{makecell}  
\usepackage{subcaption} 
\usepackage{epstopdf}
\usepackage{boldline}
\usepackage{multirow} 
\usepackage{booktabs}
\usepackage{bbm}
\usepackage[utf8]{inputenc}
\usepackage{algorithm}
\usepackage{amsmath}
\usepackage{graphicx}
\usepackage{wrapfig}
\usepackage[utf8]{inputenc}

\title{MLVTG: Mamba-Based Feature Alignment and LLM-Driven Purification for Multi-Modal Video Temporal Grounding}
{\name{Zhiyi Zhu\textsuperscript{1}, Xiaoyu Wu\textsuperscript{1}, Zihao Liu\textsuperscript{1}, Linlin Yang\textsuperscript{1}}}
{\address{\textsuperscript{1}State Key Laboratory of Media Convergence and Communication, Communication University of China}}
\begin{document}
\ninept
\maketitle
\begin{abstract}
Video Temporal Grounding (VTG), which aims to localize video clips corresponding to natural language queries, is a fundamental yet challenging task in video understanding. Existing Transformer-based methods often suffer from redundant attention and suboptimal multi-modal alignment. To address these limitations, we propose MLVTG, a novel framework comprising two key designed modules: MambaAligner and LLMRefiner. MambaAligner uses the bidirectional scanning and gated filtering strategy to model temporal dependencies and extract robust video representations for multi-modal alignment. LLMRefiner leverages the specific frozen layer of a pre-trained Large Language Model (LLM) to implicitly transfer semantic priors, enhancing multi-modal alignment without fine-tuning. This dual alignment strategy, temporal modeling via structured state-space dynamics and semantic purification via textual priors, enables more precise localization. Extensive experiments on QVHighlights, Charades-STA, and TVSum demonstrate that MLVTG achieves highly competitive performance.
\end{abstract}

\section{Introduction}
\label{sec:intro}


In video understanding, untrimmed videos dominate digital content but contain sparse valuable clips, creating strong demand for Video Temporal Grounding (VTG)—a task that aligns natural language queries with specific temporal clips in videos. VTG encompasses two core tasks: Temporal Localization (TL) and Highlight Detection (HD). However, effectively performing these tasks remains challenging due to the heterogeneous representations between visual dynamics and linguistic structures, which lead to multimodal matching ambiguities and significantly impair performance~\cite{DBLP:conf/iccv/LinZCPGWYS23}.

Early methods rely on 3D CNNs~\cite{DBLP:conf/iccv/HendricksWSSDR17,DBLP:journals/corr/abs-1907-12763,DBLP:conf/eccv/LeiYBB20}, but limit receptive fields resulted in insufficient video representation capability, leading to coarse multi-modal alignment. Transformer-based models~\cite{DBLP:conf/eccv/CarionMSUKZ20,DBLP:journals/corr/abs-2107-09609,DBLP:conf/cvpr/MoonHPPH23,DBLP:conf/iccv/LinZCPGWYS23}, leveraging the attention mechanism, have advanced VTG by effectively capturing global contextual dependencies. However, Transformers often suffer from redundant attention across frames~\cite{zeng2022not,DBLP:journals/corr/abs-2004-05150,DBLP:conf/iccv/Arnab0H0LS21}, weakening temporal discrimination. Recently, state space models(SSMs) based Mamba~\cite{DBLP:conf/iclr/GuGR22} have demonstrated strong sequence modeling capacity in vision tasks~\cite{DBLP:conf/nips/LiuTZYX0YJ024,DBLP:conf/icml/ZhuL0W0W24} by dynamically filtering redundant signals. Yet, directly applying Mamba to VTG lacks temporal expressiveness. To address this, we adopt the bidirectional scanning and the gated filtering strategy from~\cite{DBLP:conf/eccv/LiLWHWWQ24,DBLP:conf/icml/ZhuL0W0W24} to extract robust video representations for multi-modal alignment.

However, enhanced visual representations alone are still insufficient for achieving the most precise semantic alignment. To elevate powerful sequence modeling capabilities into advanced semantic comprehension and reasoning, we further introduce large language mode (LLM) to enhance multi-modal alignment. It exhibits strong multi-modal reasoning abilities~\cite{DBLP:journals/corr/abs-2401-06805,DBLP:conf/icml/DriessXSLCIWTVY23,DBLP:journals/corr/abs-2304-14178}, effectively suppressing noise, purifying semantics and enhancing multi-modal alignment in video tasks~\cite{DBLP:journals/corr/abs-2407-14500,DBLP:journals/corr/abs-2305-13292}.  To leverage such capabilities, we utilize the specific frozen layer of pre-trained LLMs. With frozen pre-trained parameters, LLMs can transfer textual priors to visual domains, filtering redundant video content. According to the Platonic Representation Hypothesis~\cite{DBLP:conf/icml/HuhC0I24}, neural networks converge to a shared semantic space across modalities, allowing LLMs to associate abstract textual concepts with temporal visual patterns, thereby improving alignment and robustness. 

\begin{figure}[!t]
  \centering
  \includegraphics[width=\linewidth]{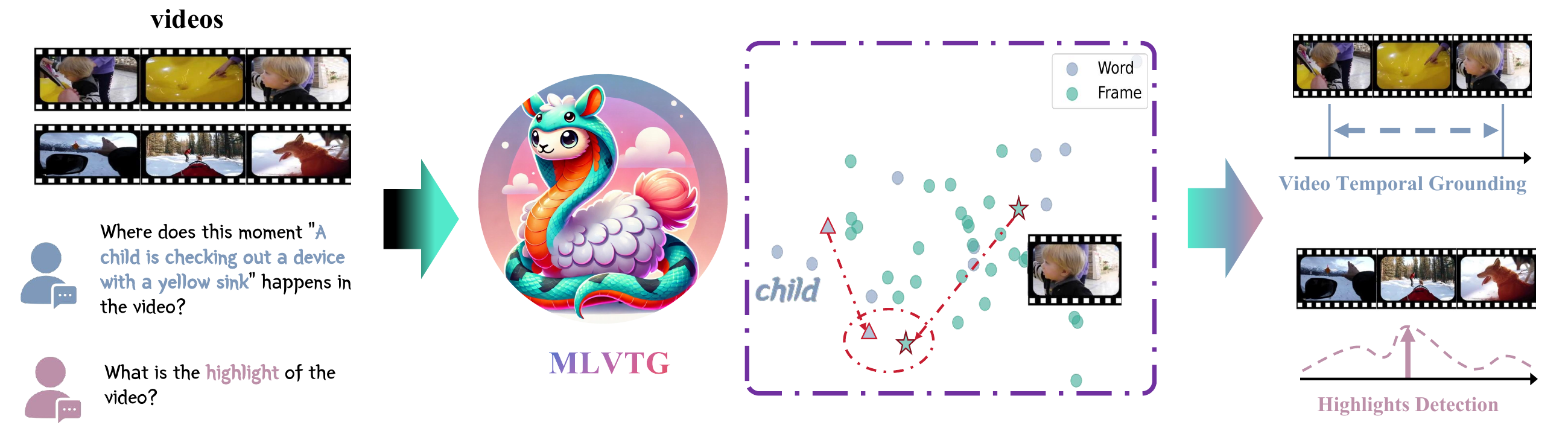}
  \caption{\textbf{MLVTG} projects features of video-text pairs into a shared semantic space, narrows the semantic gap, and better aligns visual-text features. Its dual-branch architecture separately handles temporal localization and highlight detection.}
  \label{fig:introduction}
  \vspace{-0.5cm}
\end{figure}

Overall, we propose \underline{M}amba-based feature alignment and \underline{L}LM-driven purification for multi-modal \underline{V}ideo \underline{T}emporal \underline{G}rounding termed MLVTG, a novel framework that presents MambaAligner and mamba-based LLMRefiner. MLVTG employs a dual-stage alignment strategy: the MambaAligner stage performs adaptive feature selection, while the LLMRefiner stage conducts semantic purification. MLVTG mitigates redundant attention, strengthens temporal modeling for better multi-modal alignment and introduces high-level semantic guidance without fine-tuning. To our knowledge, MLVTG is the first work to jointly exploit Mamba and LLM for VTG, achieving highly competitive performance on three benchmarks. Our key contributions include:
\begin{itemize}
    \item We propose MLVTG, a novel dual-stage framework that handles temporal localization and highlight detection separately.
    \item We propose two key modules: MambaAligner, which uses bidirectional scanning and gated filtering for robust video representation and multi-modal alignment, and a mamba-based LLMRefiner that leverages frozen LLM layers to purify semantics through implicit cross-modal interactions without fine-tuning.
    \item Extensive experiments demonstrate that MLVTG achieves highly competitive performance across three benchmarks: QVHighlights, Charades-STA, and TVSum.
\end{itemize}

\section{Related Work}
\label{sec:related}
\subsection{Video Temporal Grounding}
Video Temporal Grounding (VTG), a key task linking visual content with semantics, has evolved into two sub-tasks: Temporal Localization (TL) and Highlight Detection (HD). TL focuses on grounding natural language queries to video clips, with early proposal-based methods~\cite{DBLP:conf/iccv/HendricksWSSDR17} suffering from low coverage, while proposal-free models~\cite{DBLP:conf/eccv/LeiYBB20,DBLP:conf/acl/ZhangSJZ20} improve efficiency via direct boundary regression. HD, in contrast, identifies salient video moments, progressing from ranking-based methods~\cite{DBLP:conf/cvpr/GygliSC16} to DETR-based detectors~\cite{DBLP:conf/cvpr/MoonHPPH23} with improved accuracy. Though traditionally studied separately, TL and HD share temporal reasoning goals. M-DETR~\cite{DBLP:journals/corr/abs-2107-09609} unified them using QVHighlights, inspiring subsequent work on audio integration~\cite{DBLP:conf/cvpr/LiuLWCSQ22}, negative sample learning
, and multi-task setups~\cite{DBLP:conf/cvpr/XiaoLLMBJY024}. However, existing methods still struggle with fine-grained multi-modal alignment. To this end, we introduce MLVTG, which jointly optimizes low-level temporal structure by MambaAligner and high-level semantics by LLMRefiner, achieving improvements in both TL and HD tasks.

\subsection{Mamba for Sequence Tasks}
State Space Models (SSMs) represent long-range sequences with linear complexity~\cite{DBLP:conf/iclr/GuGR22,DBLP:conf/nips/0001GB22}, offering an efficient alternative to Transformers by leveraging continuous state representations. However, traditional SSMs lack the ability to dynamically focus on task-relevant features across varying contexts. To address this, Mamba~\cite{DBLP:journals/corr/abs-2312-00752} introduces gated selective SSMs, improving both noise suppression and salient temporal pattern extraction. Originally surpassing Transformers in language modeling and time-series forecasting~\cite{DBLP:journals/corr/abs-2312-00752,DBLP:conf/iclr/FuDSTRR23,DBLP:journals/corr/abs-2404-15772}, Mamba has recently been extended to vision task ~\cite{DBLP:conf/icml/ZhuL0W0W24,DBLP:journals/tgrs/ShenXCDY25} and video understanding~\cite{DBLP:conf/eccv/LiLWHWWQ24,DBLP:conf/nips/PanZCCZX24,DBLP:conf/nips/LiuTZYX0YJ024}. However, existing Mamba-based models primarily focus on unimodal tasks and overlook fine-grained video-language alignment. To overcome this, we incorporate a bidirectional SSM mechanism inspired by VideoMamba, and stack multiple Vision Mamba blocks to capture robust video representations. This enables precise localization for VTG tasks by enhancing temporal modeling and multi-modal alignment.

\subsection{LLM for Video Temporal Grounding}
Research is increasingly focused on applying LLMs to video tasks, exploring various methods. For example, VTimeLLM~\cite{DBLP:conf/cvpr/Huang0CS024} employs LLM as the VTG decoder to directly output start and end timestamps. However, these existing methods face significant challenges, including limited capabilities in fine-grained temporal reasoning and precise timestamp localization inherent to current LLM architectures. Additionally, the requirement for extensive training exacerbates computational overhead. Motivated by~\cite{DBLP:conf/iclr/PangXMW24}
, we propose the mamba-based LLMRefiner, a module that efficiently exploits semantic priors embedded within the specific frozen LLM layer. This strategy achieves enhanced semantic alignment between linguistic concepts and corresponding video clips, thereby significantly improving VTG performance.

\begin{figure}[!t]
    \centering
    \includegraphics[width=\linewidth]{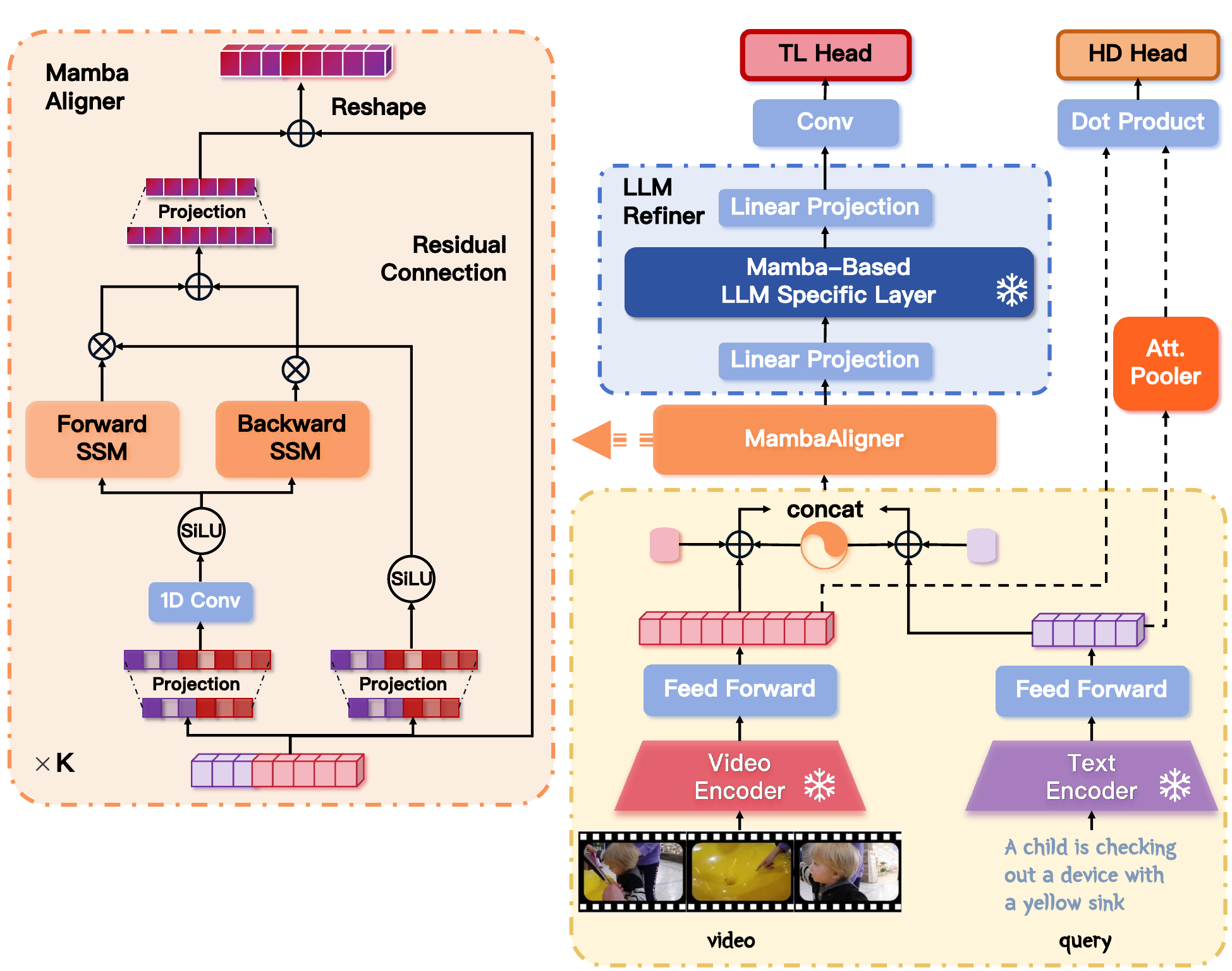}
    \caption{\textbf{Overview of the MLVTG Framework.} The MLVTG first processes the input video and query using frozen feature encoders and projects them into a shared semantic space (Sec.\ref{sec:feature_extraction}). It then computes saliency scores directly in one branch, while in the other, fused features are processed by the \textbf{MambaAligner} (Sec.\ref{sec:vision_mamba}). The aligned features undergo semantic purification via the Mamba-based \textbf{LLMRefiner} (Sec.\ref{sec:llm_refinement}). Finally, task-specific heads (Sec.\ref{sec:prediction}) classify the refined features to produce results.}
    \label{fig:framework_img}
    \vspace{-0.5cm}
\end{figure}

\begin{figure*}[!h]
\vspace{-0.5cm}
\begin{minipage}[b]{1.0\textwidth}
    \captionsetup{type=table}
    \caption{\textbf{Performances on QVHighlights.} The optimal results are bold, and the suboptimal results are underlined.}
    \renewcommand{\arraystretch}{0.9}
    \label{table_QVHighlight}
    \resizebox{\textwidth}{!}{
    \begin{tabular}{l|ccccccc|ccccccc}
    \hlineB{2.5}
    \multicolumn{1}{c|}{\textbf{Split}} & \multicolumn{7}{c|}{\textbf{Test}} & \multicolumn{7}{c}{\textbf{Val}} \\
    \cmidrule(r){1-1} \cmidrule(lr){2-8} \cmidrule(lr){9-15}
    \multicolumn{1}{c|}{\multirow{3}{*}{\textbf{Method}}} &  \multicolumn{5}{c}{\textbf{TL}}     & \multicolumn{2}{c|}{\textbf{HD}} & \multicolumn{5}{c}{\textbf{TL}}     & \multicolumn{2}{c}{\textbf{HD}} \\ 
    \cmidrule(lr){2-6} \cmidrule(lr){7-8} \cmidrule(lr){9-13} \cmidrule(lr){14-15}
    \multicolumn{1}{c|}{} & \multicolumn{2}{c}{\textbf{R1}} & \multicolumn{3}{c}{\textbf{mAP}}  & \multicolumn{2}{c|}{$\geq$ \textbf{Very Good}} & \multicolumn{2}{c}{\textbf{R1}} & \multicolumn{3}{c}{\textbf{mAP}}  & \multicolumn{2}{c}{$\geq$ \textbf{Very Good}} \\
    \multicolumn{1}{c|}{} & @0.5 & @0.7 & @0.5 & @0.75 & Avg. & mAP & HIT@1 & @0.5 & @0.7 & @0.5 & @0.75 & Avg. & mAP & HIT@1\\
    \cmidrule(r){1-1} \cmidrule(lr){2-8} \cmidrule(lr){9-15}
    M-DETR~\cite{DBLP:journals/corr/abs-2107-09609} & 52.9 & 33.0 & 54.8 & 29.4 & 30.7 & 35.7 & 55.6 & 53.9 & 34.8 & - & - & 32.2 & 35.7 & 55.6 \\
    UMT~\cite{DBLP:conf/cvpr/LiuLWCSQ22} & 56.2 & 41.2 & 53.4 & 37.0 & 36.1 & 38.2 & 60.0 & 60.3 & 44.3 & - & - & 38.6 & 39.9 & 64.2 \\
    UniVTG~\cite{DBLP:conf/iccv/LinZCPGWYS23} & 58.9 & 40.9 & 57.6 & 35.6 & 35.5 & 38.2 & 61.0 & 59.7 & - & - & - & 36.1 & 38.8 & 61.8 \\ 
    MH-DETR~\cite{DBLP:conf/ijcnn/XuSZJD24} & 60.0 & 42.4 & 60.7 & 38.1 & 38.3 & 38.2 & 60.5 & 60.8 & 44.9 & 60.7 & 39.6 & 39.2 & 38.7 & 61.7 \\
    QD-DETR~\cite{DBLP:conf/cvpr/MoonHPPH23} & 62.4 & 45.0 & 62.5 & 39.9 & 39.9 & 38.9 & 62.4 & 62.7 & 46.7 & 62.2 & 41.8 & 41.2 & 39.1 & 63.0 \\ 
    TR-DETR~\cite{DBLP:conf/aaai/SunZ0X24} & \textbf{64.6} & \textbf{48.9} & \textbf{63.9} & \underline{43.7} & 42.6 & \textbf{39.9} & 63.4 & - & - & - & - & - & - & - \\
    TaskWeave~\cite{DBLP:conf/cvpr/YangWLR24} & - & - & - & - & - & - & \textbf{64.2} & \underline{64.2} & 50.0 & \textbf{65.3} & \underline{46.4} & \underline{45.3} & \underline{39.2} & \underline{63.6} \\
    UVCOM~\cite{DBLP:conf/cvpr/XiaoLLMBJY024} & \underline{63.5} & 47.4 & \underline{63.3} & 42.6 & \textbf{43.1} & \underline{39.7} & \textbf{64.2} & 65.1 & \textbf{51.8} & - & - & \textbf{45.7} & - & - \\
    TimeChat~\cite{DBLP:conf/cvpr/RenYL0H24} & - & - & - & - & - & - & - & - & - & - & - & - & {14.5} & {23.9} \\
    VTG-LLM~\cite{DBLP:conf/aaai/GuoLLC0SLCZ25} & - & - & - & - & - & - & - & - & - & - & - & - & {16.5} & {33.5} \\
    \hlineB{2.5}
    \rowcolor{gray!20}
    \textbf{MLVTG} & {64.0} & \underline{48.3} & \underline{63.3} & \textbf{43.9} & \underline{43.0} & \textbf{39.9} & \underline{63.9}& \textbf{65.1} & \underline{50.5} & \underline{63.7} & \textbf{46.8} & {44.6} & \textbf{40.5} & \textbf{65.2} \\ 
    \hlineB{2.5}
    \end{tabular}}
    \centerline{}\medskip
\end{minipage}
\begin{minipage}[]{0.34\textwidth}
    \captionsetup{type=table}
    \renewcommand{\arraystretch}{0.8}
    \caption{\textbf{Temporal Localization Performances on Charades-STA.}}
    \label{table_charades}
    \scriptsize
    \resizebox{\linewidth}{!}{%
    \begin{tabular}{lccc}
        \toprule
         & \multicolumn{3}{l}{\textbf{Temporal Localization}} \\
        \multirow{-2}{*}{\textbf{Method}} & R1@0.5 & R1@0.7 & \multirow{-1}{*}{\textbf{mIoU}} \\ \midrule
        {M-DETR}~\cite{DBLP:journals/corr/abs-2107-09609} & 53.6 & 31.3 & - \\
        {UniVTG}~\cite{DBLP:conf/iccv/LinZCPGWYS23}  & 58.0 & 35.6 & \underline{50.1} \\
        {QD-DETR}~\cite{DBLP:conf/cvpr/MoonHPPH23}  & 55.5 & 34.1 & - \\
        {TR-DETR}~\cite{DBLP:conf/aaai/SunZ0X24}  & 57.6 & 33.5 & - \\
        {TaskWeave}~\cite{DBLP:conf/cvpr/YangWLR24}  & 56.5 & 33.6 & - \\
        {UVCOM}~\cite{DBLP:conf/cvpr/XiaoLLMBJY024} & \textbf{59.2} & \underline{36.6} & - \\
        TimeChat~\cite{DBLP:conf/cvpr/RenYL0H24} & 32.2 & 13.4 & - \\
        VTG-LLM~\cite{DBLP:conf/aaai/GuoLLC0SLCZ25} & 33.8 & 15.7 & - \\
        VTimeLLM~\cite{DBLP:conf/cvpr/Huang0CS024} & 34.3 & 14.7 & - \\
        \midrule
        \rowcolor{gray!20}
        \textbf{MLVTG} & \underline{58.3} & \textbf{38.7} & \textbf{50.3}\\
        \bottomrule
    \end{tabular}}
    \end{minipage}
\hfill
\begin{minipage}[]{0.65\textwidth}
    \captionsetup{type=table}
    \renewcommand{\arraystretch}{0.8}
    \caption{\textbf{Highlight Detection performance of Top-5 mAP on TVSum.} ``\dag'' denotes utilizing the audio modality. The suboptimal results are underlined.}
    \label{table_tvsum}
    \resizebox{\linewidth}{!}{%
    \begin{tabular}{lcccccccccc|c}
        \toprule
        \multirow{2}{*}{\textbf{Method}} & \multicolumn{11}{c}{\textbf{Highlight Detection}} \\
         & \textbf{VT} & \textbf{VU} & \textbf{GA} & \textbf{MS} & \textbf{PK} & \textbf{PR} & \textbf{FM} & \textbf{BK} & \textbf{BT} & \textbf{DS} & \textbf{Avg.} \\ \midrule
        Trailer \cite{DBLP:conf/eccv/WangLPM20} & 61.3 & 54.6 & 65.7 & 60.8 & 59.1 & 70.1 & 58.2 & 64.7 & 65.6 & 68.1 & 62.8 \\
        SL-Module \cite{DBLP:conf/iccv/XuWNZSW21} & 86.5 & 68.7 & 74.9 & \textbf{86.2} & 79.0 & 63.2 & 58.9 & 72.6 & 78.9 & 64.0 & 73.3 \\
        PLD-VHD \cite{DBLP:conf/cvpr/WeiWGJLD22} & 84.5 & 80.9 & 70.3 & 72.5 & 76.4 & \underline{87.2} & \underline{71.9} & 74.0 & 74.4 & \textbf{79.1} & 77.1 \\
        UniVTG \cite{DBLP:conf/iccv/LinZCPGWYS23} & 83.9 & \underline{85.1} & \textbf{89.0} & 80.1 & \textbf{84.6} & 81.4 & 70.9 & \textbf{91.7} & 73.5 & 69.3 & \textbf{81.0} \\
        MINI-Net\textsuperscript{\dag}~\cite{DBLP:conf/eccv/HongHLZ20} & 80.6 & 68.3 & 78.2 & 81.8 & 78.1 & 65.8 & 57.8 & 75.0 & 80.2 & 65.5 & 73.2 \\
        TCG\textsuperscript{\dag}~\cite{DBLP:conf/iccv/YeSGWBL021} & \underline{85.0} & 71.4 & 81.9 & 78.6 & 80.2 & 75.5 & 71.6 & 77.3 & 78.6 & 68.1 & 76.8 \\
        Joint-VA\textsuperscript{\dag}~\cite{DBLP:conf/iccv/BadamdorjRWC21} & 83.7 & 57.3 & 78.5 & \underline{86.1} & 80.1 & 69.2 & 70.0 & 73.0 & \textbf{97.4} & 67.5 & 76.3 \\
        CO-AV\textsuperscript{\dag}~\cite{DBLP:conf/bmvc/LiZYLLHY22} & \textbf{90.8} & 72.8 & \underline{84.6} & 85.0 & 78.3 & 78.0 & \textbf{72.8} & 77.1 & \underline{89.5} & 72.3 & \underline{80.1} \\ \midrule
        \rowcolor{gray!20}
        \textbf{MLVTG} & 83.6 & \textbf{85.6} & 74.7 & 75.9 & \underline{82.4} & \textbf{87.7} & 64.6 & \underline{91.6} & 80.9 & \underline{73.7} & \underline{80.1} \\ \bottomrule
        \end{tabular}}
    \end{minipage}
\vspace{-0.5cm}
\end{figure*}

\section{Methodology}
\label{sec:method}

As illustrated in Fig.\ref{fig:framework_img}, MLVTG presents the MambaAligner with LLMRefiner, optimizing a multi-modal feature alignment mechanism. Through a dual-branch decoupled task architecture, collaborative reasoning optimization is achieved for temporal localization and highlight detection.

\subsection{Feature Extraction and Projection.}
\label{sec:feature_extraction}
Given an input video with $L_v$ clips and a text query with $L_q$ tokens, we follow the encoder design of~\cite{DBLP:conf/iccv/LinZCPGWYS23} to extract features, which are then projected into a shared $D$-dimensional space via separate Feed-Forward Networks. This yields video features $\mathbf{V} = \{\mathbf{v}_i\}_{i=1}^{L_v} \in \mathbb{R}^{L_v \times D}$ and text features $\mathbf{Q} = \{\mathbf{q}_j\}_{j=1}^{L_q} \in \mathbb{R}^{L_q \times D}$. 

We propose a dual branch task decoupled architecture. In the one branch, an attentive pooling mechanism aggregates query tokens $\mathbf{Q} \in \mathbb{R}^{L_q \times D}$ into a sentence-level representation $\mathbf{S} \in \mathbb{R}^{1 \times D}$ to compute a saliency score via similarity measurement:
\begin{equation}
\mathbf{S} = \mathbf{M}\mathbf{Q},\quad \mathbf{M} = \text{Softmax}(\mathbf{W}\mathbf{Q}) \in \mathbb{R}^{1 \times L_q},
\end{equation}
where $\mathbf{W} \in \mathbb{R}^{1 \times L_q}$ is a learnable embedding. In another branch, we incorporate positional embeddings $\mathbf{E}^{pos}$ and modality-type embeddings $\mathbf{E}^{type}$ into both video and query tokens to preserve spatial and modality distinctions:
\begin{equation}
\tilde{\mathbf{V}} = \mathbf{V} + \mathbf{E}^{pos}_V + \mathbf{E}^{type}_V, \quad \tilde{\mathbf{Q}} = \mathbf{Q} + \mathbf{E}^{pos}_T + \mathbf{E}^{type}_T.
\end{equation}

These representations are concatenated as $\mathbf{Z} = [\tilde{\mathbf{Q}};\tilde{\mathbf{V}}] \in \mathbb{R}^{L \times D}$, where $L=L_q + L_v$. Subsequently, $\mathbf{Z}$ passes through MambaAligner for robust video representations and multi-modal alignment.

\subsection{MambaAligner.}
\label{sec:vision_mamba}
To better capture video representations for multi-modal alignment, MambaAligner consists of $K$ stacked vision mamba blocks and takes the concatenated feature sequence $\mathbf{Z} = [\tilde{\mathbf{Q}}; \tilde{\mathbf{V}}]$ as the input (see Fig.~\ref{fig:framework_img}). The input is first normalized and projected into two branches:
\begin{equation}
\hat{\mathbf{Z}} = \mathrm{Norm}(\mathbf{Z}), \quad \mathbf{x} = W_x \hat{\mathbf{Z}}, \quad \mathbf{g} = W_g \hat{\mathbf{Z}},
\end{equation}
where $\mathrm{Norm}$ denotes layer normalization, and $W_x$, $W_g$ are learnable projections. Here, $\mathbf{x}$ is used for sequence modeling, while $\mathbf{g}$ serves as the gating signal.

A 1D convolution layer extracts local features from $\mathbf{x}$, generating forward and backward sequences $\mathbf{x}^f$ and $\mathbf{x}^b$, respectively. Each is passed through a SSM with parameters $(A, B, C)$:
\begin{equation}
\mathbf{y} = \mathbf{x} * \mathbf{K}, \quad \mathbf{K} = \begin{bmatrix} 
C A^0 B, C A^1 B, \cdots, C A^{L-1} B 
\end{bmatrix},
\end{equation}
where $\mathbf{K}$ is the impulse response kernel and $*$ denotes convolution. This yields $\mathbf{y}^f$ and $\mathbf{y}^b$, capturing global context in both directions.

To fuse outputs, a gating mechanism uses the control signal $\mathbf{g}$:
\begin{equation}
\mathbf{y}_{\text{fused}} = \sigma(\mathbf{g}) \odot \mathbf{y}^f + \big(1 - \sigma(\mathbf{g})\big) \odot \mathbf{y}^b,
\end{equation}
where $\sigma(\cdot)$ is the SiLU activation, and $\odot$ denotes element-wise multiplication. The final output is computed via residual connection:
\[
\mathbf{Z}_{\text{out}} = \mathbf{Z} + \mathbf{y}_{\text{fused}}.
\]

\begin{figure*}[!h]
\vspace{-0.5cm}
  \begin{minipage}[]{0.66\textwidth}
  \captionsetup{type=table}
    \centering
    \caption{\textbf{Ablation Study on Three Benchmarks.} We select the most representative indicators as metrics for each task.}
    \label{Ablation}
    \renewcommand{\arraystretch}{0.75}
    \resizebox{\textwidth}{!}{
    \begin{tabular}{ccccccccccc}
    \toprule
    \multicolumn{1}{c}{\multirow{4}{*}{\makecell{\textbf{Mamba} \\ \textbf{Aligner}}}} & 
    \multicolumn{1}{c}{\multirow{4}{*}{\makecell{\textbf{LLM} \\ \textbf{Refiner}}}} & 
    \multicolumn{6}{c}{\textbf{Temporal Localization}} & 
    \multicolumn{3}{c}{\textbf{Highlight Detection}} \\
    \cmidrule(lr){3-8} \cmidrule(lr){9-11}
    \multicolumn{2}{c}{} & 
    \multicolumn{3}{c}{\textbf{QVHighlights}} &
    \multicolumn{3}{c}{\textbf{Charades-STA}} &
    \multicolumn{2}{c}{\textbf{QVHighlights}} &
    \multicolumn{1}{c}{\textbf{TVSum}}\\
    \cmidrule(lr){3-5} \cmidrule(lr){6-8} \cmidrule(lr){9-10} \cmidrule(lr){11-11}
    \multicolumn{2}{c}{} & 
    \multicolumn{1}{c}{\textbf{R1}} & 
    \multicolumn{1}{c}{\multirow{2}{*}{\textbf{mIoU}}} & 
    \multicolumn{1}{c}{\textbf{mAP}} & 
    \multicolumn{1}{c}{\textbf{R1}} & 
    \multicolumn{1}{c}{\multirow{2}{*}{\textbf{mIoU}}} & 
    \multicolumn{1}{c}{\textbf{mAP}} & 
    \multicolumn{2}{c}{$\geq$ \textbf{Very Good}} &
    \multicolumn{1}{c}{\multirow{2}{*}{\textbf{Avg.}}}\\
    \cmidrule(lr){5-5} \cmidrule(lr){8-8} \cmidrule(lr){9-10} 
    \multicolumn{2}{c}{} & 
    {@0.7} & {} & {Avg.} & {@0.7} & {} & {Avg.} & \textbf{HIT@1} & \textbf{mAP} \\
    \midrule
    \color{red}\usym{2718} & \color{red}\usym{2718}
     & 43.5 & 54.7 & 39.0 & 35.6 & 49.4 & 39.0 & 64.2 & 39.3 & 72.7\\
    \color{green}\usym{2714} & \color{red}\usym{2718} & {48.9} & {59.1} & {42.4} & {38.7} & {50.2} & {40.4} & {64.8} & {40.1} & 77.1\\
    \color{red}\usym{2718} & \color{green}\usym{2714} & 45.1 & 55.8 & 39.3 & 35.4 & 49.4 & 39.4 & 62.8 & 38.2 & {77.7}\\
    \rowcolor{gray!20} 
    \color{green}\usym{2714} & \color{green}\usym{2714} & \textbf{50.5} & \textbf{59.9} & \textbf{44.6} & \textbf{38.7} & \textbf{50.3} & \textbf{40.8} & \textbf{65.2} & \textbf{40.5} & \textbf{80.1}\\
    \bottomrule
    \end{tabular}}
  \end{minipage}
  \hfill
  \begin{minipage}[]{0.32\textwidth}
    \captionsetup{type=table}
    \caption{\textbf{Effectiveness of LLM Pretrained Parameters.} ``\textbf{--}'' means no LLM layer. ``{Rand}'' means randomly initialized.}
    \label{parameters}
    \scriptsize
    \resizebox{\linewidth}{!}{%
    \begin{tabular}{llcccc}
    \toprule
    \multirow{2}{*}{\makecell{\textbf{LLM} \\ \textbf{Param}}} & 
    \multirow{2}{*}{\textbf{Frozen}} & 
    \multicolumn{1}{c}{\textbf{R1}} & 
    \multicolumn{2}{c}{\textbf{mAP}} \\
    &  & {@0.7} & {@0.75} & {@Avg.} \\
    \midrule
    \textbf{--}    & \textbf{--}   & 43.5   & 39.0 & 37.7 \\
    {Rand}   & {No}   & 43.6  & 37.0 & 36.7 \\
    {Rand} & {Yes}  & 43.3  & 38.1 & 37.1 \\
    {Yes}  & {No}   & 42.8   & 36.9 & 36.4 \\
    {Yes}  & {Yes}  & \textbf{45.1} &  \textbf{39.3} & \textbf{38.4} \\
    \bottomrule
    \end{tabular}}
  \end{minipage}
\vspace{-0.5cm}
\end{figure*}

\subsection{LLMRefiner.}\label{sec:llm_refinement}
To refine the semantics of latent representations and enhance multi-modal alignment, we propose the Mamba-based LLMRefiner module. This design addresses the performance deviations caused by structural differences. The module integrates two linear layers $F_L^1$ and $F_L^2$, and a pre-trained Mamba-based large language model layer $F_{LLM}$ to improve semantic representation and multi-modal alignment. Leveraging the inherent strong multi-modal reasoning capabilities of large language models, the frozen-parameter $F_{LLM}$ layer effectively transfers textual priors into the visual domain, filtering out redundant or irrelevant information in video content.

Due to the dimension mismatch between the output of the MambaAligner($\mathbf{Z}{out}$), and the input dimension requirements of the $F_{LLM}$ layer, we integrate linear projection layers $F_L^1$ and $F_L^2$ before and after the $F_{LLM}$ block, respectively. These projection layers harmonize the dimensional compatibility, modifying the network output according to the following formulation:
\begin{equation}
    \mathbf{Z}_{refine} = F_L^1 * F_{LLM}(\mathbf{Z}_{out}) * F_L^2.
\end{equation}

In our implementation, the parameters of the pre-trained $F_{LLM}$ block remain entirely frozen to preserve their robust semantic representation capability. In contrast, the linear projection layers, $F_L^1$ and $F_L^2$, are fully trainable and optimized throughout the training process. This targeted training approach ensures continuous refinement of semantic alignment specifically tailored for downstream multi-modal alignment tasks.

\begin{figure}[!b]
\vspace{-0.5cm}
  \begin{minipage}[]{0.55\linewidth}
    \captionsetup{type=figure}
    \includegraphics[width=\linewidth]{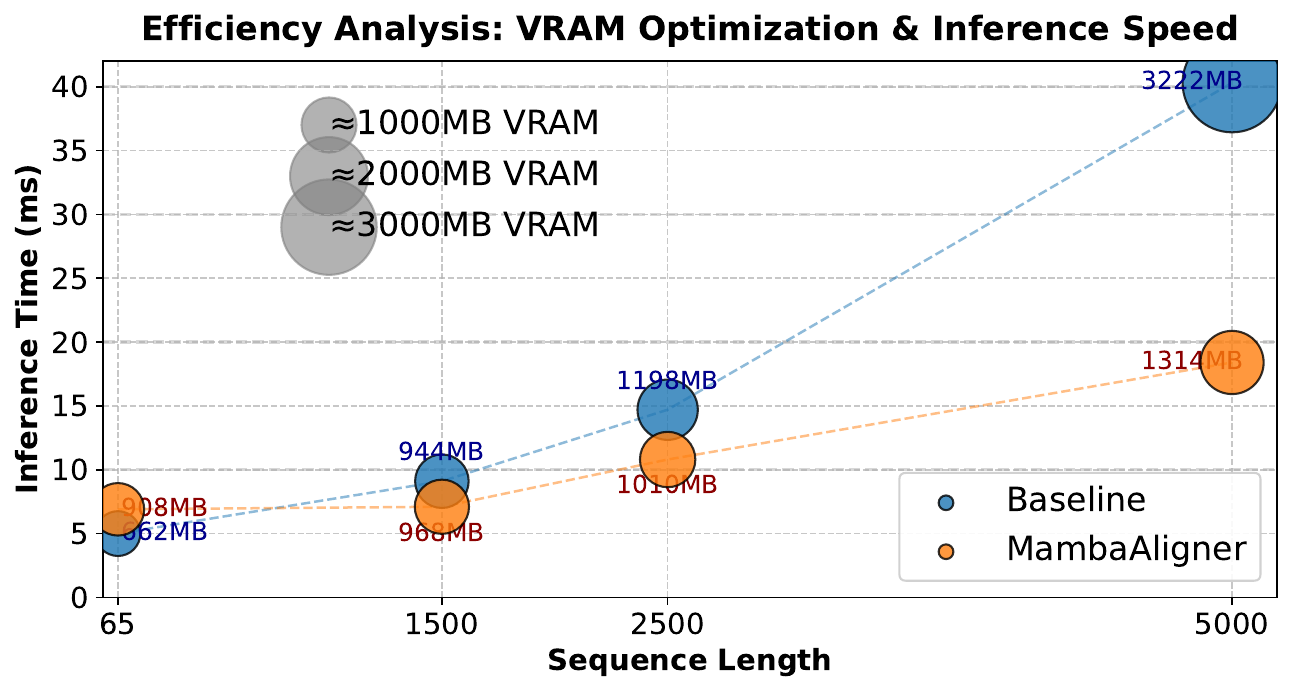}
    \captionof{figure}{\textbf{Efficiency Analysis of MambaAligner.}}
    \label{fig:efficiency_analysis}
  \end{minipage}
  \hfill
  \begin{minipage}[]{0.4\linewidth}
    \captionsetup{type=figure}
    \includegraphics[width=\linewidth]{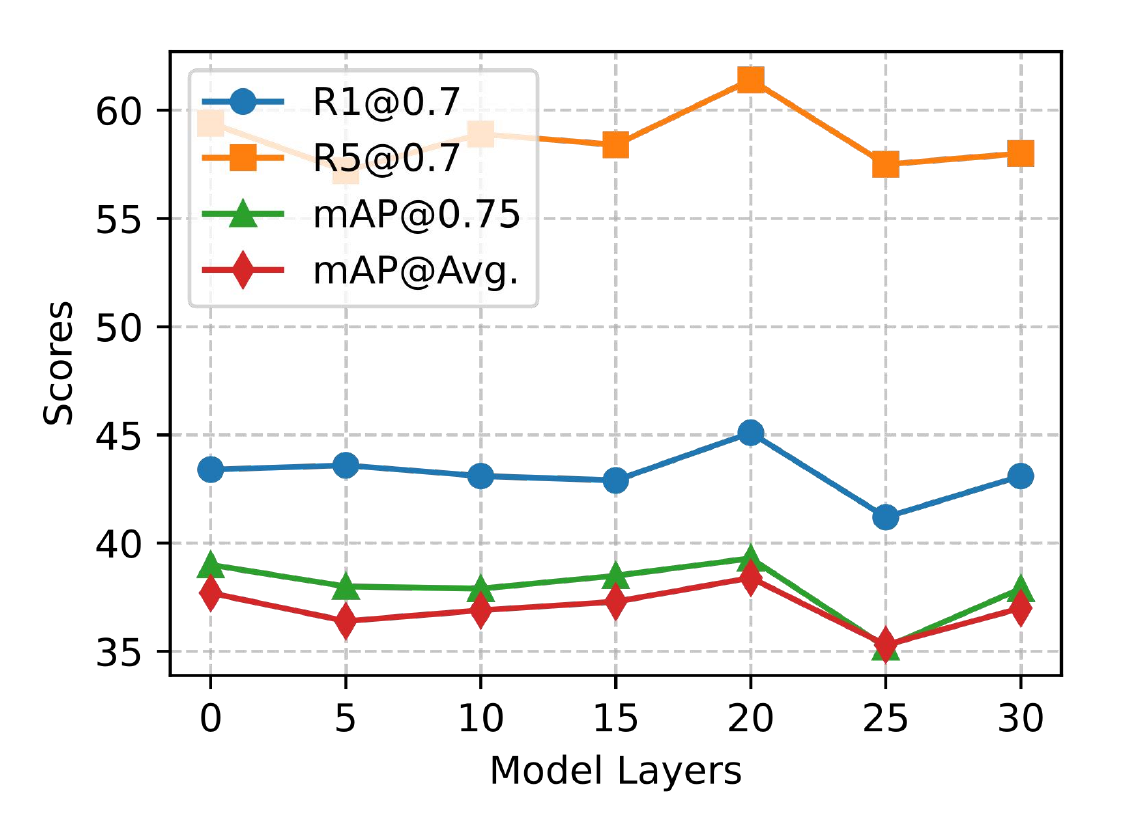}
    \captionof{figure}{\textbf{Effectiveness of LLM Different Layers.}}
    \label{fig:LLM_Layer}
  \end{minipage}
  \hfill
  \begin{minipage}[]{\linewidth}
  \captionsetup{type=figure}
    \includegraphics[width=1.0\linewidth]{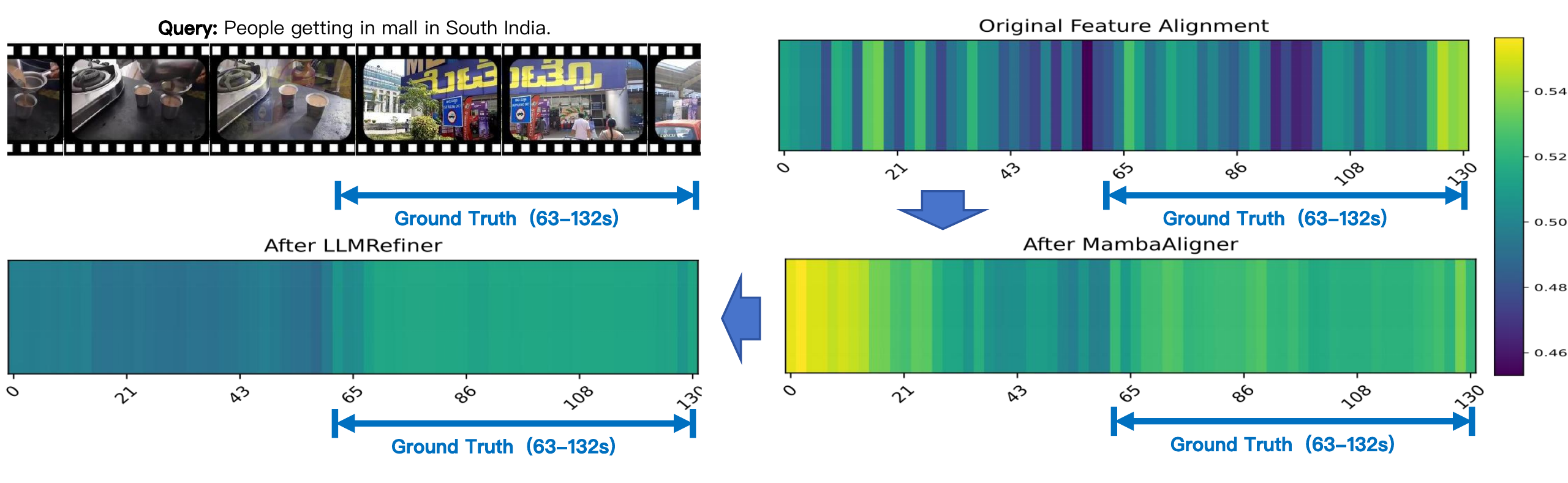}
    \caption{\textbf{Feature Alignment Visualization.} Darker colors indicate lower similarity.} \label{visualization_MLVTG}
    \end{minipage}
\end{figure}

\vspace{-0.1cm}
\subsection{Joint Optimization for TL \& HD.}\label{sec:prediction}
To jointly solve TL and HD, we design a dual-head module to produce task-specific outputs coupled with corresponding optimization objectives. Specifically, the {TL Head} takes refined feature $\mathbf{Z}_{refine}$ as input and integrates two parallel parts composed of 1D conv layers. One predicts the start and end timestamps ($[st, ed]$), while the other performs frame-level binary classification to discriminate foreground from background frames. In contrast, the {HD Head} receives video features $\mathbf{V}$ and sentence-level representations $\mathbf{S}$ as input and focuses on frame-level salient scores. The overall training objective aggregates task-specific losses with balancing coefficients, formulated as follows:

\vspace{-0.4cm}
\setlength{\jot}{0.01pt}
\begin{equation}
\begin{aligned}
    \mathcal{L}_{overall} &= \lambda_{f} \mathcal{L}_{f} + \lambda_{reg} \mathcal{L}_{reg} + \lambda_{1} \mathcal{L}^{inter} + \lambda_{2} \mathcal{L}^{intra},
\end{aligned}
\end{equation}

where $\mathcal{L}_{f}$ represents the classification loss, and $\mathcal{L}_{reg}$ denotes the regression loss, both dedicated to optimizing TL. Conversely, $\mathcal{L}^{inter}$ and $\mathcal{L}^{intra}$ are explicitly employed to enhance the accuracy and discriminative power of HD~\cite{DBLP:conf/iccv/LinZCPGWYS23}.

\section{Experiments}
\subsection{Implementation Details.} 
We use CLIP and SlowFast features from pre-segmented video clips at fixed frame rates (0.5/1 FPS), while text features are extracted through the CLIP text encoder.  The model employs a MambaAligner (4 blocks, hidden size 1024) and an LLMRefiner which operates on 2056 dimension features extracted from the 20th layer. Training used a batch size of 32 for 200 epochs on a RTX4090 GPU.

\subsection{Comparison with Start-of-the-Arts}
Table~\ref{table_QVHighlight} demonstrates outstanding performance on QVHighlights. On the test set, it achieves the highest mAP@0.75 and HD mAP. Since TR-DETR specifically designs a HD module to assist temporal localization, it performs well. However, more generalized MLVTG outperforms it on the generic TL dataset Charades-STA, leading QD-DETR by 8.6\% in R1@0.7 and outperforming both UVCOM and QD-DETR in HD task HIT@1. On the Charades-STA, MLVTG achieves optimal results under R1@0.7 and mIoU in Table~\ref{table_charades}. Moreover, without relying on complex detection architectures, it delivers superior overall performance compared to detector-based methods such as TR-DETR. On TVSum , MLVTG attains an average AP of 80.1,matching SOTA CO-AV in Table~\ref{table_tvsum}. Even without multi-dataset training like UniVTG, MLVTG remains highly competitive. And without introducing audio modality, it exhibits stronger generalization capability than complex models such as Joint-VA.

\section{Ablation Study}
As shown in the Table~\ref{Ablation}, MambaAligner alone significantly improves temporal localization performance: on QVHighlights, R1@0.7 increased by 12.4\%, mIoU by 8.0\%, and mAP@Avg by 8.7\%; on Charades-STA, it improved by 8.7\%, 1.6\%, and 3.6\%, respectively. LLMRefiner provides improvements (e.g., +3.7\% in R1@0.7 on QVHighlights), and the combination of both yields the best results, with total gains of 16.1\% in R1@0.7, 9.5\% in mIoU, and 14.4\% in mAP@Avg. On the HD task, the full model also achieves the highest performance (+3.1\% in mAP and +10.2\% in Avg). Furthermore, Table~\ref{parameters} demonstrate that using frozen pre-trained LLM parameters delivers the best outcomes (e.g., 45.1 R1@0.7 and 38.4 mAP@Avg), while fine-tuning or random initialization leads to degradation.

Fig~\ref{fig:efficiency_analysis} presents that MambaAligner demonstrates significantly lower VRAM usage and inference time compared to the baseline with longer sequence length. Although current video datasets consist of relatively short sequences, which limits full demonstration of its linear complexity advantage, the design exhibits strong potential for high efficiency in long-sequence processing. Fig~\ref{fig:LLM_Layer} further reveals that using the frozen parameters from the 20th layer performs optimally, with layers 15–22 forming a "semantic sweet spot"; either higher or lower layers result in decreased performance.

\section{Visualization}
As shown in Figure~\ref{visualization_MLVTG}, the cosine similarity heatmap between query and video features reveals poor initial alignment. After introducing MambaAligner, a substantial improvement in alignment quality is observed, though spurious high-similarity regions remain—particularly in the early clips due to background distractors. Further refinement with LLMRefiner effectively suppresses noise and concentrates high similarity within the ground-truth interval (63–132 seconds), resulting in more precise alignment.

\section{Conclusion}
This paper proposes MLVTG, a novel framework for video temporal grounding that presents MambaAligner and LLMRefiner for effective multi-modal alignment. Experiments on three benchmarks show that MLVTG outperforms highly competitive performance in both temporal localization and highlight detection. Ablation studies further confirm the effectiveness of each component, demonstrating the robustness and superiority of MLVTG. Future work will explore extending MLVTG to incorporate audio modalities for more comprehensive video temporal grounding.

\begingroup
\setstretch{0.9} 
\bibliographystyle{IEEEbib}
\bibliography{strings,refs}

@inproceedings{DBLP:conf/acl/ZhangSJZ20,
  author       = {Hao Zhang et al.},
  title        = {Span-based Localizing Network for Natural Language Video Localization},
  booktitle    = {{ACL}},
  pages        = {6543--6554},
  year         = {2020},
}

@inproceedings{DBLP:journals/corr/abs-2107-09609,
  author       = {Jie Lei et al.},
  title        = {Detecting Moments and Highlights in Videos via Natural Language Queries},
  booktitle    = {NeurIPS},
  pages        = {11846--11858},
  year         = {2021},
}

@inproceedings{DBLP:conf/cvpr/MoonHPPH23,
  author       = {WonJun Moon et al.},
  title        = {Query - Dependent Video Representation for Moment Retrieval and Highlight
                  Detection},
  booktitle    = {{CVPR}},
  pages        = {23023--23033},
  year         = {2023},
}

@inproceedings{DBLP:conf/iccv/LinZCPGWYS23,
  author       = {Kevin Qinghong Lin et al.},
  title        = {UniVTG: Towards Unified Video-Language Temporal Grounding},
  booktitle    = {{ICCV}},
  pages        = {2782--2792},
  year         = {2023},
}

@inproceedings{zeng2022not,
  title={Not all tokens are equal: Human-centric visual analysis via token clustering transformer},
  author={Zeng, Wang et al.},
  booktitle={CVPR},
  pages={11101--11111},
  year={2022}
}

@article{DBLP:journals/corr/abs-2004-05150,
  author       = {Iz Beltagy et al.},
  title        = {Longformer: The Long-Document Transformer},
  journal      = {CoRR},
  volume       = {abs/2004.05150},
  year         = {2020},
}

@inproceedings{DBLP:conf/iccv/Arnab0H0LS21,
  author       = {Anurag Arnab et al.},
  title        = {ViViT: {A} Video Vision Transformer},
  booktitle    = {{ICCV}},
  pages        = {6816--6826},
  year         = {2021},
}

@inproceedings{DBLP:conf/eccv/LiLWHWWQ24,
  author       = {Kunchang Li et al.},
  title        = {VideoMamba: State Space Model for Efficient Video Understanding},
  booktitle    = {{ECCV}},
  series       = {Lecture Notes in Computer Science},
  volume       = {15084},
  pages        = {237--255},
  year         = {2024},
}

@inproceedings{DBLP:conf/iclr/GuGR22,
  author       = {Albert Gu et al.},
  title        = {Efficiently Modeling Long Sequences with Structured State Spaces},
  booktitle    = {{ICLR}},
  year         = {2022},
}

@article{DBLP:journals/corr/abs-2401-06805,
  author       = {Yiqi Wang et al.},
  title        = {Exploring the Reasoning Abilities of Multimodal Large Language Models
                  (MLLMs): {A} Comprehensive Survey on Emerging Trends in Multimodal
                  Reasoning},
  journal      = {CoRR},
  volume       = {abs/2401.06805},
  year         = {2024},
}

@article{DBLP:journals/corr/abs-2304-14178,
  author       = {Qinghao Ye et al.},
  title        = {mPLUG-Owl: Modularization Empowers Large Language Models with Multimodality},
  journal      = {CoRR},
  volume       = {abs/2304.14178},
  year         = {2023},
}

@inproceedings{DBLP:conf/icml/DriessXSLCIWTVY23,
  author       = {Danny Driess et al.},
  title        = {PaLM-E: An Embodied Multimodal Language Model},
  booktitle    = {International Conference on Machine Learning, {ICML}},
  series       = {Proceedings of Machine Learning Research},
  volume       = {202},
  pages        = {8469--8488},
  year         = {2023},
}

@article{DBLP:journals/corr/abs-2407-14500,
  author       = {Rongkun Zheng et al.},
  title        = {ViLLa: Video Reasoning Segmentation with Large Language Model},
  journal      = {CoRR},
  volume       = {abs/2407.14500},
  year         = {2024},
}

@article{DBLP:journals/corr/abs-2305-13292,
  author       = {Guo Chen et al.},
  title        = {VideoLLM: Modeling Video Sequence with Large Language Models},
  journal      = {CoRR},
  volume       = {abs/2305.13292},
  year         = {2023},
}

@inproceedings{DBLP:conf/iccv/HendricksWSSDR17,
  author       = {Lisa Anne Hendricks et al.},
  title        = {Localizing Moments in Video with Natural Language},
  booktitle    = {{ICCV}},
  pages        = {5804--5813},
  year         = {2017},
}

@inproceedings{DBLP:conf/cvpr/GygliSC16,
  author       = {Michael Gygli et al.},
  title        = {Video2GIF: Automatic Generation of Animated GIFs from Video},
  booktitle    = {{CVPR}},
  pages        = {1001--1009},
  year         = {2016},
}

@inproceedings{DBLP:conf/cvpr/LiuLWCSQ22,
  author       = {Ye Liu et al.},
  title        = {{UMT:} Unified Multi-modal Transformers for Joint Video Moment Retrieval
                  and Highlight Detection},
  booktitle    = {{CVPR}},
  pages        = {3032--3041},
  year         = {2022},
}

@inproceedings{DBLP:conf/nips/0001GB22,
  author       = {Ankit Gupta et al.},
  title        = {Diagonal State Spaces are as Effective as Structured State Spaces},
  booktitle    = {NeurIPS},
  year         = {2022},
}

@article{DBLP:journals/corr/abs-2312-00752,
  author       = {Albert Gu et al.},
  title        = {Mamba: Linear-Time Sequence Modeling with Selective State Spaces},
  journal      = {CoRR},
  volume       = {abs/2312.00752},
  year         = {2023},
}

@inproceedings{DBLP:conf/iclr/FuDSTRR23,
  author       = {Daniel Y. Fu et al.},
  title        = {Hungry Hungry Hippos: Towards Language Modeling with State Space Models},
  booktitle    = {{ICLR}},
  year         = {2023},
}

@inproceedings{DBLP:conf/icml/ZhuL0W0W24,
  author       = {Lianghui Zhu et al.},
  title        = {Vision Mamba: Efficient Visual Representation Learning with Bidirectional
                  State Space Model},
  booktitle    = {{ICML}},
  year         = {2024},
}

@inproceedings{DBLP:conf/eccv/LeiYBB20,
  author       = {Jie Lei et al.},
  title        = {{TVR:} {A} Large-Scale Dataset for Video-Subtitle Moment Retrieval},
  booktitle    = {{ECCV}},
  series       = {Lecture Notes in Computer Science},
  volume       = {12366},
  pages        = {447--463},
  year         = {2020},
}

@inproceedings{DBLP:conf/icml/HuhC0I24,
  author       = {Minyoung Huh et al.},
  title        = {Position: The Platonic Representation Hypothesis},
  booktitle    = {{ICML}},
  year         = {2024},
}

@inproceedings{DBLP:conf/iclr/PangXMW24,
  author       = {Ziqi Pang et al.},
  title        = {Frozen Transformers in Language Models Are Effective Visual Encoder
                  Layers},
  booktitle    = {{ICLR}},
  year         = {2024},
 }

@inproceedings{DBLP:conf/eccv/CarionMSUKZ20,
  author       = {Nicolas Carion et al.},
  title        = {End-to-End Object Detection with Transformers},
  booktitle    = {{ECCV}},
  volume       = {12346},
  pages        = {213--229},
  year         = {2020},
}

@inproceedings{DBLP:conf/cvpr/Huang0CS024,
  author       = {Bin Huang et al.},
  title        = {VTimeLLM: Empower {LLM} to Grasp Video Moments},
  booktitle    = {{CVPR}},
  pages        = {14271--14280},
  year         = {2024},
}

@article{DBLP:journals/corr/abs-1907-12763,
  author       = {Victor Escorcia et al.},
  title        = {Temporal Localization of Moments in Video Collections with Natural
                  Language},
  journal      = {CoRR},
  volume       = {abs/1907.12763},
  year         = {2019},
  eprinttype    = {arXiv},
  eprint       = {1907.12763}
}

@inproceedings{DBLP:conf/ijcnn/XuSZJD24,
  author       = {Yifang Xu et al.},
  title        = {{MH-DETR:} Video Moment and Highlight Detection with Cross-modal Transformer},
  booktitle    = {{IJCNN}},
  pages        = {1--8},
  year         = {2024},
}

@inproceedings{DBLP:conf/aaai/SunZ0X24,
  author       = {Hao Sun et al.},
  title        = {{TR-DETR:} Task-Reciprocal Transformer for Joint Moment Retrieval
                  and Highlight Detection},
  booktitle    = {{AAAI}},
  pages        = {4998--5007},
  year         = {2024},
}

@inproceedings{DBLP:conf/cvpr/YangWLR24,
  author       = {Jin Yang et al.},
  title        = {Task-Driven Exploration: Decoupling and Inter-Task Feedback for Joint
                  Moment Retrieval and Highlight Detection},
  booktitle    = {{CVPR}},
  pages        = {18308--18318},
  year         = {2024},
}

@inproceedings{DBLP:conf/cvpr/XiaoLLMBJY024,
  author       = {Yicheng Xiao et al.},
  title        = {Bridging the Gap: {A} Unified Video Comprehension Framework for Moment
                  Retrieval and Highlight Detection},
  booktitle    = {{CVPR}},
  pages        = {18709--18719},
  year         = {2024},
}

@inproceedings{DBLP:conf/eccv/WangLPM20,
  author       = {Lezi Wang et al.},
  title        = {Learning Trailer Moments in Full-Length Movies with Co-Contrastive
                  Attention},
  booktitle    = {{ECCV}},
  series       = {Lecture Notes in Computer Science},
  volume       = {12363},
  pages        = {300--316},
  year         = {2020},
}

@inproceedings{DBLP:conf/iccv/XuWNZSW21,
  author       = {Minghao Xu et al.},
  title        = {Cross-category Video Highlight Detection via Set-based Learning},
  booktitle    = {{ICCV}},
  pages        = {7950--7959},
  year         = {2021},
}

@inproceedings{DBLP:conf/cvpr/WeiWGJLD22,
  author       = {Fanyue Wei et al.},
  title        = {Learning Pixel-Level Distinctions for Video Highlight Detection},
  booktitle    = {{CVPR}},
  pages        = {3063--3072},
  year         = {2022},
}

@inproceedings{DBLP:conf/eccv/HongHLZ20,
  author       = {Fa{-}Ting Hong et al.},
  title        = {MINI-Net: Multiple Instance Ranking Network for Video Highlight Detection},
  booktitle    = {{ECCV}},
  series       = {Lecture Notes in Computer Science},
  volume       = {12358},
  pages        = {345--360},
  year         = {2020},
}

@inproceedings{DBLP:conf/iccv/YeSGWBL021,
  author       = {Qinghao Ye et al.},
  title        = {Temporal Cue Guided Video Highlight Detection with Low-Rank Audio-Visual
                  Fusion},
  booktitle    = {{ICCV}},
  pages        = {7930--7939},
  year         = {2021},
}

@inproceedings{DBLP:conf/iccv/BadamdorjRWC21,
  author       = {Taivanbat Badamdorj et al.},
  title        = {Joint Visual and Audio Learning for Video Highlight Detection},
  booktitle    = {{ICCV}},
  pages        = {8107--8117},
  year         = {2021},
}

@inproceedings{DBLP:conf/bmvc/LiZYLLHY22,
  author       = {Shuaicheng Li et al.},
  title        = {Probing Visual-Audio Representation for Video Highlight Detection
                  via Hard-Pairs Guided Contrastive Learning},
  booktitle    = {{BMVC}},
  pages        = {709},
  year         = {2022},
}

@inproceedings{DBLP:conf/nips/LiuTZYX0YJ024,
  author       = {Yue Liu et al.},
  title        = {VMamba: Visual State Space Model},
  booktitle    = {NeurIPS},
  year         = {2024},
}

@article{DBLP:journals/corr/abs-2404-15772,
  author       = {Aobo Liang et al.},
  title        = {Bi-Mamba4TS: Bidirectional Mamba for Time Series Forecasting},
  journal      = {CoRR},
  volume       = {abs/2404.15772},
  year         = {2024},
}

@article{DBLP:journals/tgrs/ShenXCDY25,
  author       = {Yu Shen et al.},
  title        = {Learning Cross-Task Features With Mamba for Remote Sensing Image Multitask
                  Prediction},
  journal      = {{IEEE} Trans. Geosci. Remote. Sens.},
  volume       = {63},
  pages        = {1--16},
  year         = {2025},
}

@inproceedings{DBLP:conf/nips/PanZCCZX24,
  author       = {Zhenghao Pan et al.},
  title        = {MambaSCI: Efficient Mamba-UNet for Quad-Bayer Patterned Video Snapshot
                  Compressive Imaging},
  booktitle    = {NeurIPS},
  year         = {2024},
}

@inproceedings{DBLP:conf/aaai/GuoLLC0SLCZ25,
  author       = {Yongxin Guo et al.},
  title        = {{VTG-LLM:} Integrating Timestamp Knowledge into Video LLMs for Enhanced
                  Video Temporal Grounding},
  booktitle    = {AAAI},
  year         = {2025},
  doi          = {10.1609/AAAI.V39I3.32341},
}

@inproceedings{DBLP:conf/cvpr/RenYL0H24,
  author       = {Shuhuai Ren et al.},
  title        = {TimeChat: {A} Time-sensitive Multimodal Large Language Model for Long
                  Video Understanding},
  booktitle    = {{CVPR}},
  pages        = {14313--14323},
  year         = {2024},
}
\endgroup

\end{document}